\documentclass[11pt]{article}

\usepackage{acl}
\usepackage{times}
\usepackage{latexsym}
\usepackage[T1]{fontenc}
\usepackage[utf8]{inputenc}
\usepackage{microtype}
\usepackage{booktabs}
\usepackage{tabularx}
\usepackage{xcolor}
\usepackage{url}
\usepackage{listings}

\lstdefinestyle{prompt}{
   basicstyle=\ttfamily\scriptsize,
   columns=fullflexible,
   keepspaces=true,
   showstringspaces=false,
   breaklines=true,
   breakatwhitespace=true,
   breakindent=0pt,
   breakautoindent=false,
   postbreak={},
   frame=single,
   rulecolor=\color{black!40},
   framexleftmargin=0pt,
   aboveskip=0.5\baselineskip,
   belowskip=0.5\baselineskip
}

\title{Task-Preserving Multilingual Instruction Tuning for Open-Ended QA in WMT 2026 MIST}

 \author{
  Aleix Sant$^{1}$\textbf{, }
  Jordi Luque$^{1}$\textbf{, }
  Carlos Escolano$^{2}$ \\[4pt]
  $^{1}$Scientific Research, Telefónica Innovación Digital \\
  $^{2}$Universitat Politècnica de Catalunya \\
  Barcelona, Spain \\[2pt]
  \texttt{\small \{aleix.santsavall,jordi.luqueserrano\}@telefonica.com} \\
  \texttt{\small carlos.escolano@upc.edu}
}

\begin{document}
\maketitle

\begin{abstract}
This paper describes our submission to the WMT 2026 Multilingual Instruction Shared Task. We adapt Qwen3-8B with LoRA using organiser-provided data, EuroAlpaca, a multilingual instruction dataset created through task-preserving localisation, and CrossEuroAlpaca, a cross-lingual augmentation that assigns different source and target languages to instructions, contexts and responses. Our system targets open-ended question answering (QA-OEG) in ten shared-task languages represented in the auxiliary data. Under matched decoding, adaptation improves the mean Gemma-4 judge score on target-language QA-OEG by 7.8\% relative to the base model. On an additional evaluation using the Aya Evaluation Suite across all shared-task languages, it yields relative gains of 41.5\% in macro-averaged ROUGE-L and 2.9\% in BERTScore-F1, with larger improvements in target languages. However, these gains are task- and language-specific. Gemma-4 scores improve for QA-OEG and summarisation in the target languages but decline for context-grounded QA there and across all three tasks in the non-target languages.

\end{abstract}

\section{Introduction}
\label{sec:introduction}

Instruction tuning fine-tunes LLMs on labelled datasets of instructional prompts and their corresponding responses. It improves performance on the represented tasks while strengthening the model's general ability to follow unseen instructions \citep{DBLP:conf/iclr/WeiBZGYLDDL22,DBLP:conf/nips/Ouyang0JAWMZASR22}. Multilingual instruction tuning extends this objective across languages and writing systems, requiring models to interpret instructions in different languages and respond in the requested language. This remains challenging because instruction resources are unevenly distributed across languages. A scalable way to expand multilingual coverage is to translate every textual field of English instruction examples directly. However, such translations may alter the underlying task or its constraints, thereby corrupting the supervision signal \citep{sant2026euroalpacataskpreservinglocalisationinstruction}. 

The WMT 2026 Multilingual Instruction Shared Task (MIST) studies this setting through summarisation and both context-grounded and open-ended question answering \citep{schmidtova-etal-2026-findings}. Inputs span multiple languages and may explicitly constrain the response language, making some instances cross-lingual. The shared task permits external training data and the use of models under 10B parameters. We focus on open-ended question answering (QA-OEG) in Czech, German, English, Finnish, Italian, Portuguese, Russian, Slovak, Spanish and Turkish. Nevertheless, we produce predictions for the complete test set to assess generalisation to other languages and to tasks that are less represented in our training mixture.

Our system applies LoRA \citep{hu2022lora} to the Qwen3-8B model. The training mixture combines the data supplied by the organisers (we refer to it as MIST), plus additional auxiliary data: EuroAlpaca, a multilingual instruction-tuning dataset created through task-preserving localisation, and CrossEuroAlpaca, a cross-lingual augmentation balanced across all directed source--target language pairs. The submitted checkpoint is selected based on validation loss, and inference uses greedy decoding with a repetition penalty chosen through a matched sensitivity analysis of the base and adapted models.

The adapted model improves the target QA-OEG judge score by 7.8\% over the identically decoded base model. Additional evaluation on the Aya Evaluation Suite also shows semantic gains across all 27 languages, with larger improvements in the languages covered by the auxiliary data. These gains do not extend uniformly across the shared-task formats. Context-grounded QA deteriorates within the target set, while scores decline across all three tasks outside the target languages. These results highlight the task- and language-specific gains and regressions introduced by the adapter.
\section{Shared Task}
\label{sec:task}

MIST evaluates three forms of multilingual instruction following. QA-context requires models to answer a question using a supplied passage, QA-OEG covers open-ended requests without a grounding passage, and SUM-SUM requires models to summarise a supplied document. The test set includes both monolingual and cross-lingual examples. QA-context may use different languages for the passage and the question or response, SUM-SUM requires summarisation of English documents into multiple target languages, and some QA-OEG prompts explicitly request output in a language different from that of the instruction. The provided test set contains 12,775 prompts across 24 test languages, comprising 8,640 QA-context, 2,359 QA-OEG and 1,776 SUM-SUM instances. We optimise our system for QA-OEG in the ten response languages covered by our auxiliary data. 

\section{Related Work}
\label{sec:related-work}


\paragraph{Multilingual Instruction Tuning.}
Multilingual instruction tuning adapts language models on instruction--response pairs spanning multiple languages and writing systems, with the aim of transferring instruction-following capabilities across languages. Large-scale resources such as xP3 and Aya demonstrate that joint multilingual training can support such cross-lingual transfer \citep{muennighoff2023crosslingual, DBLP:conf/acl/SinghVD0MKSPMOZ24, DBLP:conf/acl/UstunAYKDOBSOKV24}. Translating English instruction datasets offers a scalable way to broaden language coverage, although the resulting models remain sensitive to the linguistic and task composition of the training data \citep{sant2026euroalpacataskpreservinglocalisationinstruction}. Controlled comparisons further show that multilingual tuning can approach or match separate monolingual adapters, even with limited supervision in individual languages \citep{chen2024monolingual, DBLP:conf/acl/0002HASTE24}.

\paragraph{Task-preserving Localisation.}
Alpaca-style instruction data organise each example into an \emph{instruction} describing the task, an optional \emph{input} providing context, and an \emph{output} representing the expected response. Translating these fields uniformly can invalidate tasks whose solutions depend on linguistic form, including spelling, grammar, rhyme, character counts, fixed labels, or source and target language roles \citep{DBLP:conf/emnlp/ChenYGH24, DBLP:conf/acl/LiYZLWZ24}. EuroAlpaca shows that these inconsistencies can reduce instruction-following performance and introduces task-preserving localisation to address them \citep{sant2026euroalpacataskpreservinglocalisationinstruction}. Its pipeline determines whether an example can be translated safely, preserves fields that must remain unchanged, and rewrites language-dependent examples so that the task remains valid and solvable in the target language. We therefore use EuroAlpaca as auxiliary supervision, favouring task-valid localised examples over corpora produced through uniform machine translation.

\paragraph{Reference-free Evaluation.}
When reference answers are unavailable, an increasingly common approach is to use a language model as a judge. The evaluator assesses each candidate response against its prompt and a predefined rubric. G-Eval and the LLM-judge protocol used in MT-Bench are prominent examples of model-based evaluation for open-ended generation \citep{DBLP:conf/emnlp/LiuIXWXZ23, DBLP:conf/nips/ZhengC00WZL0LXZ23}. This setting is relevant to our submission because participants have neither gold responses for the test prompts nor scores from a private held-out evaluation set to guide model selection. However, model judgements may reflect evaluator preferences and uneven calibration across languages \citep{hada2024evaluators, fu2025multilingualjudge}. We therefore treat judge scores as diagnostic evidence, evaluating every candidate with the same model, rubric and test instances, and complementing this analysis with reference-based metrics from the Aya Evaluation Suite.
\section{Training and Validation Data}
\label{sec:data}

\begin{table}[t]
\centering
\small
\begin{tabular}{lrr}
\toprule
\textbf{Source} & \textbf{Original} & \textbf{Effective} \\
\midrule
EuroAlpaca & 569,360 & 569,360 \\
CrossEuroAlpaca & 569,360 & 569,360 \\
MIST (cleaned) & 21,444 & 85,776 \\
\midrule
Total & 1,160,164 & 1,224,496 \\
\bottomrule
\end{tabular}
\caption{Training data.}
\label{tab:data-mixture}
\end{table}

Table~\ref{tab:data-mixture} summarises the final training mixture. We create a deterministic 90/10 split of the MIST data, stratified by language, task and source dataset. After some cleaning, this yields 21,444 training examples and 2,384 validation examples. We repeat the MIST training split four times to reduce its dilution by the substantially larger auxiliary corpora and maintain alignment with the shared-task data distribution. The final training set combines these repeated MIST examples with EuroAlpaca and CrossEuroAlpaca, followed by deterministic shuffling. The held-out MIST samples are used exclusively for validation and early stopping.

We convert each record into a single user--assistant turn. When the input field is empty, the user message contains only the instruction. When an input is present, the user message contains the instruction followed by a blank line and the input. The record's output forms the assistant response. We then render the turn using Qwen's native chat template.

\subsection{MIST}
We apply lightweight cleaning to the organiser data. We remove three QA rows whose reference duplicates the prompt and 13 occurrences of exact duplicates, leaving 21,444 training examples. The 2,384 held-out validation examples remain separate. We repeat the cleaned training set four times, yielding 85,776 effective samples within the much larger auxiliary mixture. By doing so, we raise its share from 1.8\% of the training mixture to 7.0\% and maintain regular exposure to supervision that directly matches the shared-task formats.

\subsection{EuroAlpaca}
We use the public EuroAlpaca data in Czech, German, English, Finnish, French, Italian, Portuguese, Russian, Slovak and Spanish. For this work, we additionally create a Turkish version using the same pipeline. Each language contains 51,760 examples, giving 569,360 examples across the 11 languages. Records follow the Alpaca-style schema of an \emph{instruction}, an optional \emph{input} and an \emph{output}.

The EuroAlpaca pipeline first predicts the task and domain, and then determines whether direct translation is safe for each example. It then assigns field-level translation or preservation decisions, protecting content whose original form is necessary for the task. Examples that cannot be translated safely are instead rewritten at the task level. A final coherence check evaluates instruction--response consistency, language correctness, grammar and mixed-language artefacts.

\subsection{CrossEuroAlpaca}
We construct CrossEuroAlpaca for this shared task by converting localised EuroAlpaca records into explicitly cross-lingual examples. Using EuroAlpaca's translation-policy annotations, we retain 40,672 underlying examples classified as safe for field-wise translation and exclude 11,088 language-dependent examples. To preserve the original task, we combine only aligned localisations of the same underlying example. When context is present, the instruction and output use the target language, while the context uses a different source language. For context-free examples, the instruction uses the source language and the output uses the target language. Each prompt includes an explicit response-language directive, such as \textit{Antworte auf Deutsch} or \textit{Türkçe cevap verin}, taken from the shared-task prompts when available. This aligns the auxiliary examples with the response-language constraints used in the shared task.

The 11 languages form 110 ordered source--target directions, excluding same-language pairs. German-to-Turkish and Turkish-to-German, for example, are separate directions. Each direction contains 5,176 examples, producing 569,360 rows in total. Every language therefore appears 51,760 times as both source and target. Assignment is deterministic, and each underlying Alpaca example occurs at most once per direction.
\section{System Description}
\label{sec:system}

\subsection{Model and Adaptation}
The base model is Qwen3-8B, a dense decoder-only transformer with 36 layers, grouped-query attention and multilingual instruction-following capabilities.\footnote{\href{https://huggingface.co/Qwen/Qwen3-8B}{Qwen3-8B model card}.} We apply LoRA to all attention query, key, value and output projections, with rank 16, $\alpha=32$ and dropout 0.05. This trains 15.3M of 8.20B parameters
, leaving the base model frozen.

\subsection{Training}
Training uses AdamW, bfloat16 precision, gradient clipping at 1.0, no weight decay and 100 linear warm-up updates from $10^{-7}$ to the peak learning rate, which is $1.5\times10^{-5}$. We use DeepSpeed data parallelism across four 64~GB NVIDIA H100 GPUs, with a micro-batch size of two per GPU and 16 gradient accumulation steps, yielding an effective batch size of 128. 
Early stopping monitors validation loss with patience 5 and a minimum improvement of 0.001. We evaluate every quarter of an epoch. 

\begin{table*}[t]
\centering
\small
\begin{tabular}{lrrr@{\qquad}lrrr}
\toprule
\textbf{Language} & \textbf{Base} & \textbf{Ours} & \textbf{$\Delta$} &
\textbf{Language} & \textbf{Base} & \textbf{Ours} & \textbf{$\Delta$} \\
\midrule
Czech      & 2.99 & 3.53 & +0.54 & Portuguese & 4.04 & 4.28 & +0.24 \\
German     & 3.92 & 4.04 & +0.12 & Russian    & 3.70 & 4.12 & +0.42 \\
English    & 4.49 & 4.35 & -0.14 & Slovak     & 2.80 & 3.26 & +0.46 \\
Finnish    & 2.65 & 2.90 & +0.25 & Spanish    & 4.10 & 4.42 & +0.32 \\
Italian    & 3.96 & 4.27 & +0.31 & Turkish    & 3.31 & 3.59 & +0.28 \\
\midrule
Mean score & 3.60 & \textbf{3.88} & +0.28 & Acceptance (\%) & 49.6 & \textbf{62.6} & +13.0 \\
\bottomrule
\end{tabular}
\caption{Gemma-4 results for QA-OEG in the ten target languages. Both systems use greedy decoding with repetition penalty 1.08. Each language has 100 instances. Acceptance is the percentage scoring four or five.}
\label{tab:target-results}
\end{table*}

\subsection{Inference}
Each shared-task prompt is passed as a single user message through Qwen's native chat template, with thinking mode disabled. We set the maximum prompt length to 16,384 tokens to accommodate the long, paper-length documents used in the summarisation task. Generation is limited to 1,024 new tokens. Our final system uses greedy decoding with a repetition penalty of 1.08. In a pilot experiment, five-beam decoding occasionally caused language and script drift and was therefore discarded. We use an inference batch size of eight.
\section{Evaluation}
\label{sec:setup}

\subsection{Shared-task Test Analysis}




We compare off-the-shelf Qwen3-8B with the LoRA-adapted model under matched inference conditions. Results are analysed by task and by whether the requested response language belongs to our ten-language\footnote{Note that French was included in the original MIST data and therefore in our EuroAlpaca-based auxiliary data, but it was not part of the shared-task test set. Consequently, we do not report Gemma-4 judge scores for French.} target set.

Because reference answers are unavailable, we use a Q8-quantised Gemma-4-31B\footnote{\href{https://huggingface.co/google/gemma-4-31B-it}{Gemma-4 model card}.} instruction model as a reference-free judge. The judge receives the complete task prompt, task and language metadata, and candidate response. It assigns an integer score from one to five based on task completion, instruction and language compliance, grounding, factuality, fluency and the absence of visible internal reasoning. We define acceptance rate as the proportion of responses receiving a score of four or five. Judge decoding uses temperature zero. Appendix~\ref{app:judge-prompt} reproduces the complete evaluation prompt.

As shown in Table~\ref{tab:decoding}, we evaluate both the base and adapted models with repetition penalties of 1.00, 1.02, 1.05, 1.08 and 1.10. We select 1.08 for the final system because it produces the highest target QA-OEG score for the adapted model. We evaluate the base model with the same repetition penalty of 1.08 so that the score difference is not caused by different decoding parameters. The repetition penalty was selected using Gemma-4 judge scores on the released test prompts, so the decoding analysis is test-aware. 


The target QA-OEG subset contains 1,000 instances, formed by instantiating the same 100 underlying prompts in each of the ten target languages. To estimate uncertainty in the paired mean difference between automatic Gemma-4 judge scores, we treat each prompt and its ten language versions as one bootstrap cluster. We report a percentile-based 95\% confidence interval obtained from 50,000 cluster resamples using seed 42.

\subsection{Aya Evaluation Suite}
As a complementary evaluation, we evaluate the base and our adapted model on the Aya Evaluation Suite, which provides multilingual open-ended tasks with references for automatic evaluation \citep{DBLP:conf/acl/SinghVD0MKSPMOZ24}. We report ROUGE-L for lexical overlap \citep{lin2004rouge} and BERTScore-F1 with XLM-RoBERTa-large for semantic similarity \citep{zhang2020bertscore}. Scores are macro-averaged over languages. We evaluate 27 languages corresponding to the overlap between the Aya Evaluation Suite and the MIST sample-data languages. Both base and adapted models use identical 8,192-token contexts, batches of 16, greedy decoding without repetition control, and at most 1,024 new tokens. For each automatic metric, we assess how sensitive the macro-average paired difference is to the set of evaluated languages using a language-level bootstrap. We resample languages with replacement and report percentile-based 95\% confidence intervals from 100,000 resamples using seed 42.
\section{Results}
\label{sec:results}

\subsection{Target QA-OEG}

Table~\ref{tab:target-results} presents the main results for target QA-OEG. Under matched greedy decoding with a repetition penalty of 1.08, the adapted model improves

\begin{table*}[t]
\centering
\small
\begin{tabular}{lrrr@{\qquad}rrr}
\toprule
& \multicolumn{3}{c}{\textbf{ROUGE-L}} & \multicolumn{3}{c}{\textbf{BERTScore-F1}} \\
\cmidrule(lr){2-4}\cmidrule(lr){5-7}
\textbf{Languages} & \textbf{Base} & \textbf{Adapted} & \textbf{$\Delta$ [95\% CI]}
& \textbf{Base} & \textbf{Adapted} & \textbf{$\Delta$ [95\% CI]} \\
\midrule
All $\rightarrow$ 27       & 13.56 & \textbf{19.20} & +5.63 [3.49, 7.81]
             & 84.51 & \textbf{87.00} & +2.49 [2.12, 2.85] \\
Adaptation $\rightarrow$ 11 & 16.90 & \textbf{27.89} & +10.99 [8.67, 12.73]
             & 85.42 & \textbf{88.86} & +3.45 [3.32, 3.59] \\
Submission $\rightarrow$ 10 & 16.68 & \textbf{27.55} & +10.88 [8.35, 12.79]
             & 85.37 & \textbf{88.83} & +3.46 [3.32, 3.61] \\
\bottomrule
\end{tabular}
\caption{Aya Evaluation Suite results, macro-averaged over languages and scaled by 100. Intervals are language-level bootstrap confidence intervals. Adaptation includes French. Submission contains the 10 languages targeted in the shared-task test set. Both models use matched greedy decoding without repetition control.}
\label{tab:aya-results}
\end{table*}

\noindent
the mean Gemma-4 judge score from 3.596 to 3.876, corresponding to a relative improvement of 7.8\%. The paired mean difference is $+0.280$, with a 95\% cluster-bootstrap confidence interval of $[+0.133,+0.429]$. The interval remains above zero under prompt-level cluster resampling, indicating that the estimated gain is stable to changes in the sampled prompts. The acceptance rate increases from 49.6\% to 62.6\%, a gain of 13.0 percentage points.

Nine of the ten target languages improve. The largest gains occur in Czech, Slovak and Russian, while English is the only language to decline. The aggregate improvement is therefore distributed across most of the target-language set rather than being concentrated in one or two languages.

Table~\ref{tab:decoding} compares the base and adapted models across the same repetition penalties. Moderate repetition control improves both systems, and the adapted model outperforms the base model at every tested value. Its target QA-OEG score peaks at 1.08 before declining at 1.10, supporting the use of 1.08 for the final system. The positive adapted-model advantage across all tested values shows that the target gain is not confined to the selected repetition penalty.

\begin{table}[t]
\centering
\small
\begin{tabular}{rrrrr}
\toprule
\textbf{RP} & \multicolumn{2}{c}{\textbf{Target QA-OEG}} & \multicolumn{2}{c}{\textbf{All tasks}} \\
\cmidrule(lr){2-3}\cmidrule(lr){4-5}
 & \textbf{Base} & \textbf{Adapted} & \textbf{Base} & \textbf{Adapted} \\
\midrule
1.00 & 3.400 & 3.510 & 3.551 & 2.995 \\
1.02 & 3.485 & 3.679 & 3.587 & 3.042 \\
1.05 & 3.533 & 3.805 & 3.591 & 3.108 \\
\textbf{1.08} & 3.596 & \textbf{3.876} & \textbf{3.614} & \textbf{3.130} \\
1.10 & \textbf{3.600} & 3.828 & 3.604 & 3.128 \\
\bottomrule
\end{tabular}
\caption{Repetition-penalty (RP) sensitivity under greedy decoding. Target QA-OEG is macro-averaged over the ten target languages.}
\label{tab:decoding}
\end{table}

\subsection{Task and Language Trade-offs}
Table~\ref{tab:task-tradeoff} extends the comparison to the complete test distribution. Within the target-language set, adaptation improves QA-OEG and summarisation but substantially weakens context-grounded QA, which exhibits the largest regression. Outside the target languages, scores decline across all three tasks. Consequently, pooling all test instances favours the base model despite the improvement on the targeted task.

\begin{table}[t]
\centering
\small
\begin{tabular}{llrrr}
\toprule
\textbf{Languages} & \textbf{Task} & \textbf{Base} & \textbf{Adapt.} & \textbf{$\Delta$} \\
\midrule
Target & QA-context & 4.178 & 3.347 & -0.831 \\
Target & QA-OEG     & 3.596 & 3.876 & +0.280 \\
Target & SUM-SUM    & 3.318 & 3.509 & +0.191 \\
\midrule
Other  & QA-context & 3.638 & 3.038 & -0.600 \\
Other  & QA-OEG     & 2.930 & 2.825 & -0.105 \\
Other  & SUM-SUM    & 2.658 & 2.234 & -0.424 \\
\midrule
All & All tasks & 3.614 & 3.130 & -0.484 \\
\bottomrule
\end{tabular}
\caption{Gemma-4 scores by task and language coverage at repetition penalty 1.08. Target denotes the ten submission target languages.}
\label{tab:task-tradeoff}
\end{table}

\subsection{Aya Evaluation Suite}

Table~\ref{tab:aya-results} presents the complementary reference-based evaluation on the Aya Evaluation Suite. BERTScore-F1 improves in all 27 evaluated languages, while ROUGE-L improves in 21. The macro-average gains are considerably larger across the 11 languages represented in the auxiliary training data and remain nearly unchanged when French is excluded to match the ten target languages. Every language in both subsets improves on both metrics.

All reported confidence intervals from the language-level bootstrap exclude zero, indicating positive macro-average differences under language resampling. The six ROUGE-L regressions occur outside the target set, and BERTScore-F1 still improves in each case. On Aya, adaptation therefore increases semantic similarity consistently, although greater lexical overlap is not observed in every language.

\subsection{Official Shared-Task Results}
\label{sec:official_results}

The official MIST evaluation assigns our submitted system an overall QA-OEG score of 62.37 across all 24 evaluation languages. On the ten target languages, the average score is 73.63. Table~\ref{tab:official_results} reports the per-language results. Our system ranks second among teams in English, Portuguese and Spanish, and third in Italian, while being less competitive in Czech and Finnish.

\begin{table}[t]
    \centering
    \small
    \begin{tabular}{lcc}
        \toprule
        \textbf{Language} & \textbf{Score} & \textbf{Team rank} \\
        \midrule
        Czech      & 61.5 & 6/7 \\
        German     & 80.3 & 4/7 \\
        English    & 91.4 & 2/7 \\
        Finnish    & 47.8 & 6/7 \\
        Italian    & 83.4 & 3/7 \\
        Portuguese & 86.6 & 2/7 \\
        Russian    & 78.3 & 4/7 \\
        Slovak     & 55.1 & 5/7 \\
        Spanish    & 86.0 & 2/7 \\
        Turkish    & 65.9 & 5/7 \\
        \midrule
        \textbf{Average} & 73.63 & -- \\
        \bottomrule
    \end{tabular}
    \caption{Official QA-OEG results on the ten target languages. Team ranks collapse multiple submissions from the same team by retaining the highest score per language.}
    \label{tab:official_results}
\end{table}

These scores are not directly comparable to the Gemma-4 results above, as the official QA-OEG metric uses a majority-vote constraint-rule pass rate across three LLM judges.
\section{Analysis}
\label{sec:discussion}
The results show selective specialisation by both task and language. Within the target-language set, adaptation improves open-ended QA and summarisation but substantially weakens context-grounded QA. Outside the target set, scores decline across all three tasks. Target-language gains therefore do not guarantee that performance will be preserved across task formats or languages not covered by the auxiliary data.

The Aya Evaluation Suite provides complementary evidence. Improvements in BERTScore-F1 across all 27 evaluated languages and in ROUGE-L across 21 indicate broader semantic and lexical similarity to reference responses. This does not contradict the context-grounded QA regression. Aya measures reference similarity on open-ended responses, whereas QA-context requires grounding in a supplied passage. The system should therefore not be characterised as either globally improved or globally degraded.

The official evaluation provides a further perspective on this language-specific behaviour. Competitive rank does not necessarily follow the gains over the base model. For example, English ranks second among teams despite a small decline under our Gemma-4 comparison. This reflects the different comparisons involved, as our analysis measures adaptation gains over the base model, whereas the official evaluation compares submitted systems under a different metric.

Training composition offers one plausible explanation. MIST accounts for only 7.0\% of the effective training rows and is the only source drawn directly from the shared-task distribution. The much larger auxiliary corpora may favour target-language response generation without providing enough task-matched supervision to preserve context-grounded QA. However, the reported experiments do not isolate the contribution of each data source, so the observed effects cannot be attributed specifically to EuroAlpaca or CrossEuroAlpaca. Since the base parameters remain frozen, the regressions are consistent with reversible adapter-induced interference rather than permanent loss of the base model's capabilities. More balanced task sampling, targeted QA-context replay and task-aware checkpoint selection could help reduce this trade-off.

\section{Conclusion}
\label{sec:conclusion}
We presented a Qwen3-8B system adapted with LoRA on MIST, EuroAlpaca and CrossEuroAlpaca, targeting open-ended QA in ten languages. Under matched decoding, LoRA adaptation improves Gemma-4 judge scores on target-language QA-OEG and summarisation in the target languages, while the Aya Evaluation Suite shows broader multilingual gains in reference-based response quality. These gains coexist with a substantial decline in context-grounded QA and lower scores across all three tasks outside the target-language set. These trade-offs highlight the need to preserve task coverage alongside language coverage during multilingual adaptation.

\section*{Limitations}
\label{sec:limitations}

Gemma-4 is a single reference-free judge, and we do not validate its scores against human annotations or a second evaluator. Its judgements may therefore reflect uneven language coverage, rubric interpretation or response-style preferences \citep{hada2024evaluators, fu2025multilingualjudge}.


\section*{Acknowledgment}
This work has received funding from the European Union's Horizon Europe research and innovation programme under the project ELOQUENCE (Grant Agreement No. 101135916). This work was supported by computational resources from the EuroHPC Joint Undertaking under the EuroHPC AI Factory grant EHPC-AIF-2026LS01-004.

\bibliography{references}

\appendix
\section{LLM-as-a-Judge Prompt}
\label{app:judge-prompt}

The following prompts were sent to Gemma-4. Placeholders in braces were replaced for each test instance.

\medskip
\noindent\textbf{System prompt}\par
\nobreak
\begin{lstlisting}[style=prompt]
Return valid JSON only. Do not include markdown.
\end{lstlisting}

\medskip
\noindent\textbf{User prompt template}\par
\nobreak
\begin{lstlisting}[style=prompt]
You are a strict multilingual evaluation judge for a shared-task test set.

Your job is to decide how well the candidate answer follows the prompt and solves the task.
There is no reference answer. Judge only from the prompt and the candidate answer.

General judging rules:
- Follow explicit instructions in the prompt first: requested language, format, length, style, constraints, and task type.
- Judge only what is supported by the prompt. Do not assume hidden information or require a reference answer.
- Penalize answers that ignore the task type (e.g., summarizing when asked to answer a question, or giving generic advice when asked to use the provided context).
- Do not reward unnecessary verbosity. A longer answer is better only if it adds correct, relevant information.
- Penalize malformed output if the prompt asks for a specific format.

Language correctness:
- If the prompt requests a specific output language, the answer must be in that language.
- If the prompt itself is in the target language, the answer should use that same language unless the prompt says otherwise.
- Penalize wrong language, wrong script, unnecessary language mixing, bad spelling, bad grammar, or clearly unnatural wording.
- Do not over-penalize minor wording issues if the answer is clearly correct, fluent, and in the right language.

Content correctness:
- Penalize missing constraints, hallucinated facts, unsupported claims, empty answers, irrelevant text, and visible internal reasoning such as <think>...</think>.
- For summarization, reward faithful coverage of the main points and penalize invented details or missing central information.
- For question answering, reward direct correctness and penalize evasive, unsupported, or overlong answers when the prompt asks for a concise answer.
- For context-based question answering, reward answers grounded in the provided context and penalize claims not supported by it.

Scoring rubric (1-5 integer scale):
5 = Fully correct, complete, fluent, follows all important instructions, and is written in the correct language.
4 = Mostly correct and in the correct language, with only minor omissions, formatting issues, or wording problems.
3 = Partially correct, but with important omissions, some incorrect content, weak grounding, or noticeable language-quality problems.
2 = Mostly wrong, incomplete, poorly grounded, not useful, or written mainly in the wrong language.
1 = Wrong, unrelated, empty, unreadable, unusable, or contains visible internal reasoning.

Set acceptable = true only for scores 4 or 5.

Return only this JSON object, with no extra text:
{"score": 1-5, "acceptable": true/false, "reason": "short explanation"}

Task metadata: {task}
Language metadata: {question_lang}

Prompt:
{prompt}

Candidate answer:
{output}
\end{lstlisting}

\end{document}